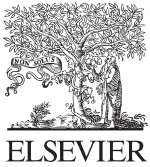

International Conference on Advanced Computing Technologies and Applications (ICACTA-2015)

# Animal Classification System: A Block Based Approach

Y H Sharath Kumar*, Manohar N, Chethan H K

*Maharaja Institute of Technology Mysore, India*
*High Performance Computing project  DoS in CS, University of Mysore, Mysore 570 006, India*
sharathyhk@gmail.com, manohar.mallik@gmail.com, hkchethan@gmail.com

**Abstract**

In this work, we propose a method for the classification of animal in images. Initially, a graph cut based method is used to perform segmentation in order to eliminate the background from the given image. The segmented animal images are partitioned in to number of blocks and then the color texture moments are extracted from different blocks. Probabilistic neural network and K-nearest neighbors are considered here for classification.  To corroborate the efficacy of the proposed method, an experiment was conducted on our own data set of 25 classes of animals, which consisted of 4000 sample images. The experiment was conducted by picking images randomly from the database to study the effect of classification accuracy, and the results show that the K-nearest neighbors classifier achieves good performance.

## 1. Introduction

For several years, detection of animals in wildlife footage is a great area of interest among biologist. Often biologists study the behaviour of animals to understand and predict the actions of animals. Also detection of animals has got several applications such as animal-vehicle accident prevention, and animal trace facility, identification, anti-theft, security of animals in zoo. Presently researchers identifying animals manually this is tedious and time consuming. Since the dataset is very large, manual identification is a daunting task. Computer-assisted animal classification makes this work efficient and reduces the time. Wildlife images captured in the field represent a challenging task in classification of animals since it appears with different pose, cluttered background, different light and climate conditions, different view point and occlusions. Additionally the animal of different classes looks similar. All these challenges need an efficient algorithm to classify the animals.

---

\* Corresponding author. Tel.: +91-9880277488. *E-mail address:* sharathyhk@gmail.com

Classification of animals mainly has three stages viz., segmentation, feature extraction, and classification. Segmentation subdivides the image into its constituent objects. The goal is to do object recognition. Since the object of our interest is some part in the image and it is combined with other regions, segmentation is carried out to get the object of our interest and discard the background. Animals are often surrounded by plants or trees and shadow in the background. So segmentation is carried out to get the animal region which is only the region of our interest in it.

Pixel based segmentation, which uses only pixel appearance to assign a label to a pixel (Das et al., [1], Tobias et al., [2]). Region based segmentation, the algorithm detects for valid segments at each scale further segments extracted at various scale are integrated to get the final result ((Susanta and Bhabatosh, [3]). In Graph-based labelling methods (Boykov and Jolly, [4], Nilsback and Zisserman, [5]), a global energy function is defined depending on both appearance and image gradients. Mean shift based segmentation is been widely-used for the segmentation which includes two steps, filtering the original image data in feature space and clustering of the filtered data points (Comaniciu D, Meer P [6]).

Once the animal object is extracted from the image the next step is feature extraction. Feature extraction involves in extracting the different properties which can effectively classify. Some animals possess high distinctive shapes, some have distinctive color, some have distinctive texture patterns, and some are characterized by a combination of these properties.

Tilo et al., [7] proposed a method for animal classification which uses facial features. Mayank et al., [8] differentiate the individual zebra by using their coat markings. Ardovini et al., [9] presented a model for the identification of elephant based on the shape of the nicks of the elephant. Deva et al., [10] built 2D articulated models and this model is used to detect the animal in the videos by using the texture feature of the animals. The animals are classified using unsupervised learning (Pooya et al., [11]). Heydar et al., [12] developed an animal classification system using joint textural information. Human/animal classification for unattended ground sensors uses wavelet statistics based on average, variance and energy of the third scale residue and spectral statistics based on amplitude and shape features for robust discrimination (Ranganarayanaswami et al., [13]). Matthias Zeppelzauer [14] focused on automated detection of elephants in wildlife video which uses color models. Xiaoyuan Yu et al., [15] proposed a method for automated identification of animal species in camera trap images. In this method they used dense SIFT descriptor and cell-structured LBP (CLBP).

After feature extraction, the challenge lies in deciding suitable classifier. Mansi et al., [16] proposed a model for Animal detection using template matching. Matthias and Zeppelzauer [14] used color features for animal classification using Support Vector Machine (SVM) classifier. Deva et al., [10] tried a variety of classifiers, such as K-way logistic regression, SVM, and K-Nearest Neighbors. Tree based classifiers are capable of modelling complex decision surfaces (Ranganarayanaswami et al., [13]). Heydar et al., [12] use three different classifiers like Single Histogram, SVM and Joint Probability model. Pooya et al., [11] present a model for animal detect by comparing with ground truth created by human experts. Xiaoyuan et al., [15] used linear SVM classifier for the classification of animals.

From the literature survey it is understood that, there are quite a few attempts towards development of a animal classification system. An also we can observe that, the majority of work is carried for small datasets. Also simple classifier such as nearest neighbors classifier, support vector machine has been used. Hence in this work we divide the animal image into blocks of size 1, 4, 16, 64 from each block the color texture moments (CTM) are extracted. The extracted features are applied to fusion classifier of probabilistic neural network and K-nearest neighbors for classification of animals.

## 2. Proposed System

In the proposed method the input animal image is segmented by a graph-cut based technique. We divide the segmented animal image in to number of blocks. We study the performance by considering entire image as a single block, and also by splitting the image into 4, 16 and 64 blocks. Color texture moments (CTM) features are extracted from each block. These features are queried to Probabilistic neural network and K-nearest neighbors, to know the class label of unknown animal.

*2.1 Segmentation*

Animals are often surrounded by greenery background in images. Hence, the extraction of only animals by eliminating the background is necessary. In order to avoid considering the features of background region rather than the desired animal region, the image need to be segmented. Currently, segmentation systems are semi-automatic where the user has to mark the foreground and background regions. In our system, the user has to mark a few regions to identify background and foreground regions; but once those regions are selected, further operations are performed by the system automatically without the need of any additional information. Rather than using the original image, use of segmented animal image brings an improvement in the system's performance. In our work, we use Iterated Graph Cuts Algorithm (Boykov et al., [4]) to segment the animal images from the background. The initial labelling $f_0$ of graph cuts is labelled by a group of foreground/background seeds from the user. Regions which have pixels marked as foreground are called foreground seed regions, while the regions with background seeds are called background seed regions. Starting from the initial sub-graph, adjacent regions to the previously labelled regions are added into the updated sub-graph iteratively. Optimal segmentation is obtained by running graph cuts algorithm on the updated sub-graph until all the region nodes are labelled as either foreground (i.e. object) or background. Fig. 1 shows examples of segmented images.

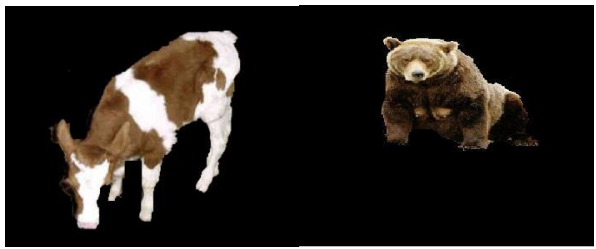

Fig.1. Example of segmented images

*2.2.1 Color space selection*

A color image is represented by three components, such as RGB, XYZ, YIQ, HIS and this color spaces play a significant role in classification of animals. In our work, we perform the feature extraction using HSV color space due to its perceptual uniformity. The HSV color space is a non-linear transform of the RGB space and is widely used in the field of computer vision. The HSV of a pixel can be transformed from its RGB representation using the following formula.

HSV=Hue, Saturation, Value
RGB= Red, Green, Blue

$$H = \arctan \frac{\sqrt{3}(G-B)}{(R-G)+(R-B)} \quad (1)$$

$$S = 1 - \frac{\min\{R,G,B\}}{V} \quad (2)$$

$$V = \frac{(R+G+B)}{3} \quad (3)$$

To overcome problem of HSV color space, we represent color by a three dimensional vector $x = (x_1, x_2, x_3)$, where

$$x_1 = S*V*\cos(H) \quad (4)$$

$$x_2 = S*V*\sin(H) \quad (5)$$

$$x_3 = V \quad (6)$$

*2.2.2 Local Fourier Transform*

We make use of the local Fourier transform to extract features for representing the local grey-tone spatial dependency. The local Fourier transform is equivalent to eight unique templates as shown in Table 1 which operate on the image respectively [17]. We utilize these eight templates to extract the local Fourier coefficients and hence eight characteristic maps are obtained.

Table 1: Eight templates for computing characteristic maps

| 1 | 1 | 1 |
|---|---|---|
| 1 | 0 | 1 |
| 1 | 1 | 1 |

| -1 | 1 | -1 |
|----|---|----|
| 1  | 0 | 1  |
| -1 | 1 | -1 |

| $-\frac{\sqrt{2}}{2}$ | 0 | $\frac{\sqrt{2}}{2}$ |
|---|---|---|
| -1 | 0 | 1 |
| $-\frac{\sqrt{2}}{2}$ | 0 | $\frac{\sqrt{2}}{2}$ |

| $-\frac{\sqrt{2}}{2}$ | -1 | $-\frac{\sqrt{2}}{2}$ |
|---|---|---|
| 0 | 0 | 0 |
| $\frac{\sqrt{2}}{2}$ | 1 | $\frac{\sqrt{2}}{2}$ |

| 0 | -1 | 0 |
|---|----|---|
| 1 | 0  | 1 |
| 0 | -1 | 0 |

| 1  | 0 | -1 |
|----|---|----|
| 0  | 0 | 0  |
| -1 | 0 | 1  |

| $\frac{\sqrt{2}}{2}$ | 0 | $-\frac{\sqrt{2}}{2}$ |
|---|---|---|
| -1 | 0 | 1 |
| $\frac{\sqrt{2}}{2}$ | 0 | $-\frac{\sqrt{2}}{2}$ |

| $-\frac{\sqrt{2}}{2}$ | 1 | $-\frac{\sqrt{2}}{2}$ |
|---|---|---|
| 0 | 0 | 0 |
| $\frac{\sqrt{2}}{2}$ | -1 | $\frac{\sqrt{2}}{2}$ |

*2.2.3 Color Feature Extraction*

The color distribution can be uniquely characterized by its moments. Most of the color distribution information can be captured by the lower-order moments. It is also proved that the first and second order moments are good approximation in representing color distributions of images. We extract the first and second order moments of each characteristic map obtained by the application of local fourier transform. For every color channel, the moments are calculated independently in each of the eight maps and then a 16-dimensional feature vector is obtained. By concatenating the feature vectors extracted in different color channels, the color information can be utilized completely. For example, when (SVcosH, SVsinH, V) color space is considered, the dimension of color texture moments (CTM) will become 48. Intuitively, we can also regard this method as an expansion of the color moments through eight feature spaces that are orthogonal to each other. Therefore, it inherits the advantages of color moments, as well as improves the characteristic capabilities.

### 3. Classification

The extracted color texture moments is fed to fusion classifier of nearest neighbor and Probabilistic Neural Network. Introductions on each of these classifiers are given in the following subsections.

*3.1 K- Nearest Neighbor (KNN)*

One of the simplest classifier amongst all the classifiers is the K-Nearest Neighbor classifier [18] [19]. The term K-nearest can be interpreted as the k-number of points which are relatively closer (with respect to a distance) to a point in n-dimensional feature space. In concern with this, the Euclidean distance is computed (in particular) between a test sample and the train samples. The k-training samples with relatively less distances are termed as K-Nearest Neighbors. So, among k-training samples, the maximum number of samples which are most similar to a test sample, proves out to allocate its class label to the given test sample. This exploits the 'smoothness' assumption that samples close to each other will probably have the same class. In our work we consider K=1.

*3.2 Probabilistic Neural Networks (PNN)*

Neural network are employed to classify patterns based on studying from examples. Different neural network paradigms employ different learning rules, but all in same way determine pattern statistics from a set of training samples and then classify new patterns on set basis of these statistics. Probabilistic neural network [20] [21] is one type of classifier built with feed forward networks, where no feedback exists in a network. The input layer is fully connected to the hidden layer which has a node for each classification. Each hidden node calculates the dot product of the input vector with a test vector subtracts one from it and divides the result by the squared standard deviation. The output layer has a node for each pattern class. The sum of each hidden node is sent to the output layer and the highest value wins. Here, the training is done in single pass rather than multiple passes. For representation of training samples, probabilistic neural network estimates the probability density function for each class and the same is calculated for each test sample. The Probabilistic neural network trains immediately but execution time is slow and it requires a large amount of space in memory.

### 4. Database

The available animal dataset show less intra class variations and do not have much change in its view point. Hence, we have created our own animal dataset by collecting images from World Wide Web and by taking photographs of animals be found in and around our place. Our dataset consists of 4000 images that are divided into 30 different classes where, each class varies from 40 to 300 images. The images are taken to study the effect of the proposed method with large intra class variations and different viewpoints. Fig. 2(a) shows sample image of different classes and Fig. 2(b) shows samples images of randomly selected animal classes with intra and inter class variations. The large intra-class variability and the small inter-class variability makes this dataset very challenging.

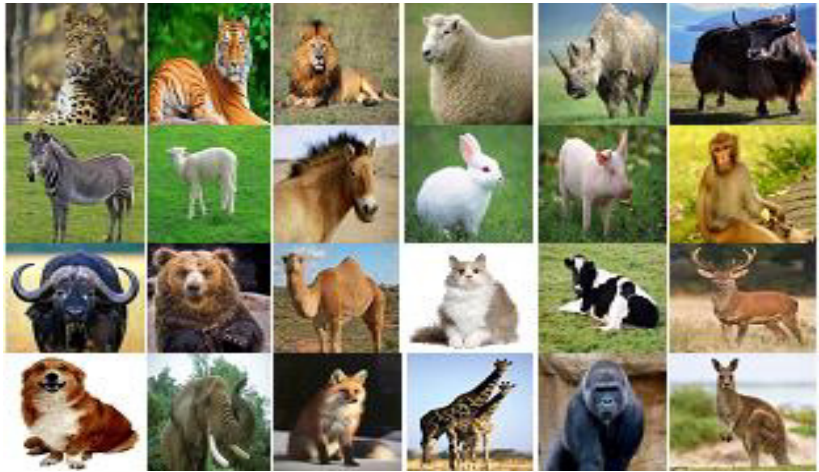

Fig. 2. (a) Sample images of different class

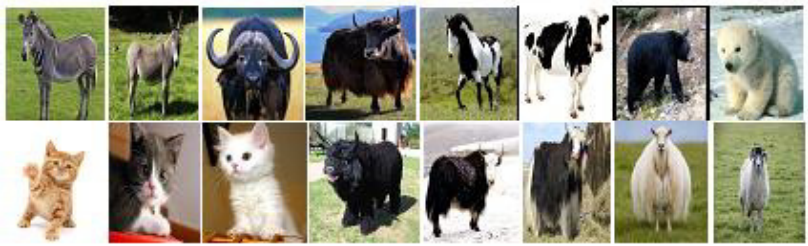

Fig. 2. (b) Sample images with intra and inter class variations

## 5. Results

In our work, we measure the performances of different classifiers. We picked images randomly from the database and experimentation is conducted more than five times by splitting images into a different number of blocks. We report maximum, minimum and average of the accuracy obtained in all cases. For all blocks the experimentation has been conducted on database of 25 classes under varying training samples from 30 to 70 percent of database. The results obtained for single block, 4, 16 and 64 blocks are respectively tabulated in Table2 and Table3. In each table, the results are tabulated for each individual classifier with varying training samples. It can be noticed that the KNN classifiers achieves relatively higher accuracy in all cases, When 70 percent of the database system are used for training.

Table 2. Accuracy of different classifier and their combinations by treating an image as single and 4 block

| Block | Training Samples (%) | Accuracy (%) | PNN | KNN | Block | Training Samples (%) | Accuracy (%) | PNN | KNN |
|---|---|---|---|---|---|---|---|---|---|
| 1 | 70 | Max | 78.09 | 82.70 | 4 | 70 | Max | 70.46 | 72.91 |
|  |  | Min | 77.52 | 81.41 |  |  | Min | 70.16 | 72.19 |
|  |  | Avg | 77.70 | 82.38 |  |  | Avg | 70.20 | 72.55 |
|  | 50 | Max | 75.68 | 79.91 |  | 50 | Max | 66.56 | 68.79 |
|  |  | Min | 73.31 | 77.83 |  |  | Min | 65.76 | 68.14 |
|  |  | Avg | 74.26 | 78.49 |  |  | Avg | 66.04 | 68.30 |
|  | 30 | Max | 69.69 | 74.07 |  | 30 | Max | 59.35 | 62.99 |
|  |  | Min | 67.11 | 71.61 |  |  | Min | 57.92 | 60.19 |
|  |  | Avg | 68.52 | 72.69 |  |  | Avg | 58.48 | 61.48 |

Table 3: Accuracy of different classifier and their combinations by treating an image as 16 and 64 block

| Block | Training Samples (%) | Accuracy (%) | PNN | KNN | Block | Training Samples (%) | Accuracy (%) | PNN | KNN |
|---|---|---|---|---|---|---|---|---|---|
| 16 | 70 | Max | 73.19 | 76.80 | 64 | 70 | Max | 60.16 | 68.87 |
|  |  | Min | 69.44 | 72.76 |  |  | Min | 52.08 | 61.73 |
|  |  | Avg | 71.97 | 74.96 |  |  | Avg | 56.70 | 65.32 |
|  | 50 | Max | 69.29 | 73.02 |  | 50 | Max | 55.71 | 66.17 |
|  |  | Min | 67.21 | 70.15 |  |  | Min | 48.17 | 60.93 |
|  |  | Avg | 68.91 | 72.09 |  |  | Avg | 54.72 | 63.71 |
|  | 30 | Max | 64.62 | 68.07 |  | 30 | Max | 54.12 | 65.91 |
|  |  | Min | 62.80 | 65.86 |  |  | Min | 42.17 | 54.81 |
|  |  | Avg | 63.40 | 67.12 |  |  | Avg | 49.69 | 61.33 |

## 6. Conclusion

We have developed a block based algorithm for animal classification. In our work the animal images are divided into different number of blocks and performance of all classifiers is showed. It is observed that the proposed KNN classifier achieves relatively good classification accuracy when compared to any other PNN classifier. We have created our own database of animals of 25 classes containing 4000 animal images. We conducted experimentation under varying number of blocks of animals and we studied its effect on classification accuracy. The experimental results have shown that the KNN classifier achieved better results when compared to any other classifiers.

**Acknowledgments**

The work by Manohar N was supported by High Performance Computing project lab, University of Mysore, Mysore.